%% file: main.tex
\documentclass[cameraready]{template/mac_automl_xmu_blue}

\makeatletter
\def\input@path{{content/}}
\makeatother
\graphicspath{{content/}}

\input{resources/packages}

\input{resources/math_macro}

\setleftheadercontent{%
  \headerlogospace{1.6mm}%
  \headerlogo[-0.08\height]{15.2mm}{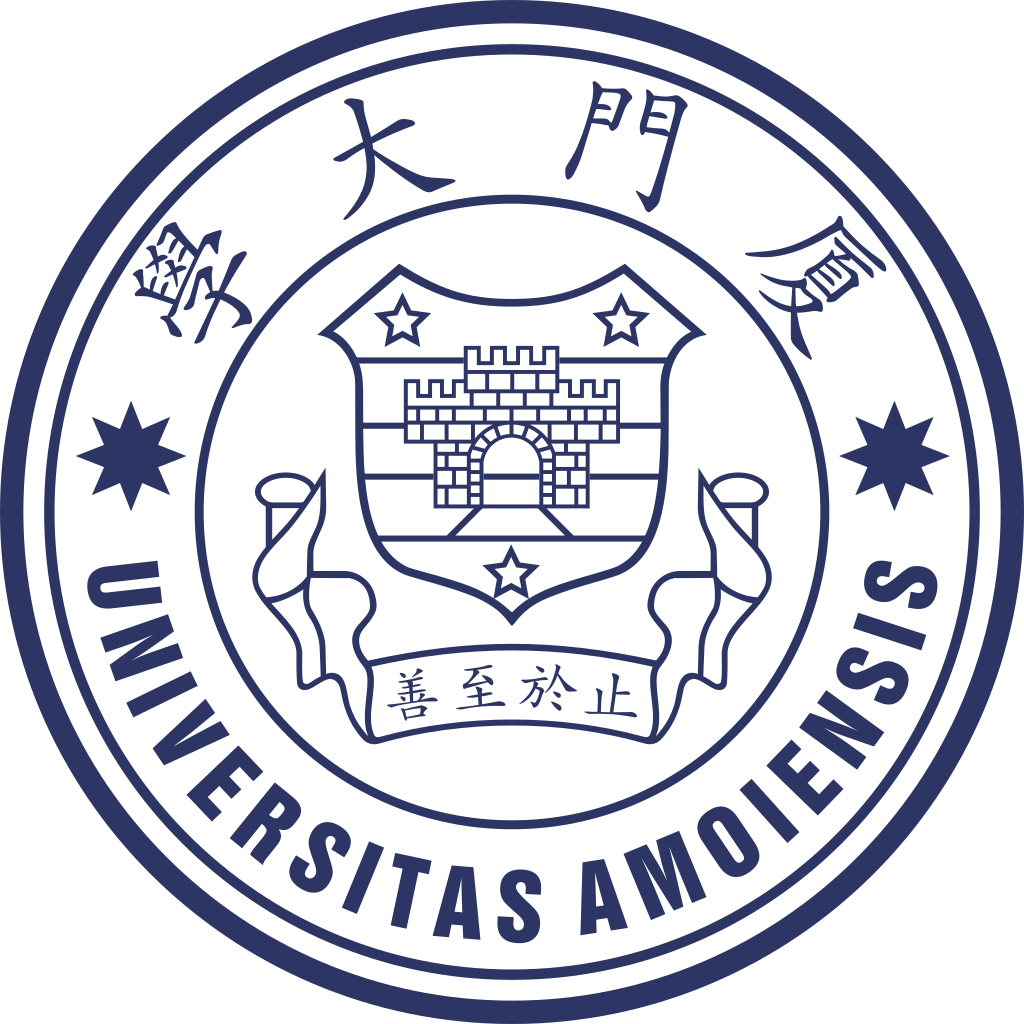}%
  \headerlogospace{2.8mm}%
  \headerlogo[-0.05\height]{12.8mm}{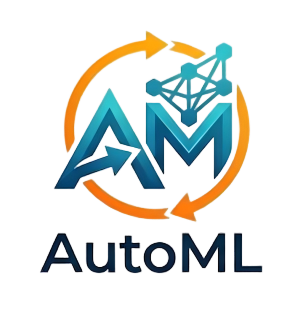}%
}
\setrightheadericon{}
\setrunningheadericon{%
  \headerlogospace{1.2mm}%
  \headerlogo[-0.03\height]{8.2mm}{assets/branding/automl.png}%
}
\setheadergroupname{MAC-AutoML}

\title{Toward Robust In-Context Segmentation\\via Concept Guidance}

\setfrontauthors{%
  \authorrow{%
    \authorentry{Zhigang Chen}{1},\authorsep
    \authorentry{Xiawu Zheng}{1},\authorsep
    \authorentry{Rongrong Ji}{1,\corremailmark}}%
}
\setfrontaffiliations{%
  \affiliationline{\authormark{1}Key Laboratory of Multimedia Trusted Perception and Efficient Computing, Ministry of Education of China, Xiamen University, 361005, P.R. China}
}
\setfrontcontact{%
  \contactline{\textbf{\corremailmark\ Corresponding Author:} Rongrong Ji}
}
\usecustomauthorlayout

\checkdata[Email]{\href{mailto:chenzg11@stu.xmu.edu.cn}{\nolinkurl{chenzg11@stu.xmu.edu.cn}}}

\abstract{\input{sec/0_abstract}}

\begin{document}

\maketitle

\input{sec/1_intro}
\input{sec/2_related}
\input{sec/3_method}
\input{sec/4_experiments}
\input{sec/5_conclusion}

\clearpage
\bibliographystyle{plainnat}
\bibliography{references}

\end{document}

%% file: content/resources/packages.tex
\usepackage{xargs}
\usepackage{todonotes}
\usepackage{tabularx}

%% file: content/resources/math_macro.tex
\newcommand{\graytext}[1]{\textcolor{gray}{#1}}

%% file: content/sec/0_abstract.tex
In-context segmentation (ICS) requires a model to segment target regions in a query image using only a few reference images and their corresponding masks, without updating any parameters. 
Despite recent progress, prior ICS studies have largely overlooked a critical aspect: \emph{system robustness}, \ie, whether the model can produce stable segmentation results for the same query under different references.  
In this work, we revisit ICS from the robustness perspective and introduce a novel paradigm, Concept-Guided In-Context Segmentation (CG-ICS), which performs segmentation by extracting high-level semantic concepts from references rather than relying solely on low-level visual matching. Specifically, CG-ICS introduces a concept reasoning module that uses an MLLM to propose candidates and a SAM3-driven scoring function with tree-search refinement to select reliable textual concepts, together with a parallel visual exemplar route that provides query-side spatial grounding via a simple context construction. Both the textual concept and the visual exemplar are then used to activate the segmentation capability of a frozen SAM3 backbone. Extensive experiments on standard ICS benchmarks demonstrate that CG-ICS not only achieves state-of-the-art accuracy but also substantially improves robustness, yielding a more reliable ICS system with significantly reduced variance across diverse reference choices. Code is available at \url{https://github.com/Kakarot1103/CG-ICS}.

%% file: content/sec/1_intro.tex
\section{Introduction}
\label{sec:intro}

Image segmentation stands as a fundamental task in computer vision. However, traditional fully supervised approaches are restricted by their heavy reliance on costly annotations and limited generalization to novel categories. Recently, inspired by the success of In-context Learning in Large Language Models (LLMs)~\cite{brown2020language}, In-Context Segmentation (ICS)~\cite{bar2022visual, painter23,seggpt23} has emerged to address these challenges. 
ICS aims to segment the target region in a query image using only a few reference examples, without updating any model parameters. It offers a flexible and efficient solution for open-world segmentation.

\begin{figure}[t]
\centering
\includegraphics[width=0.95\textwidth]{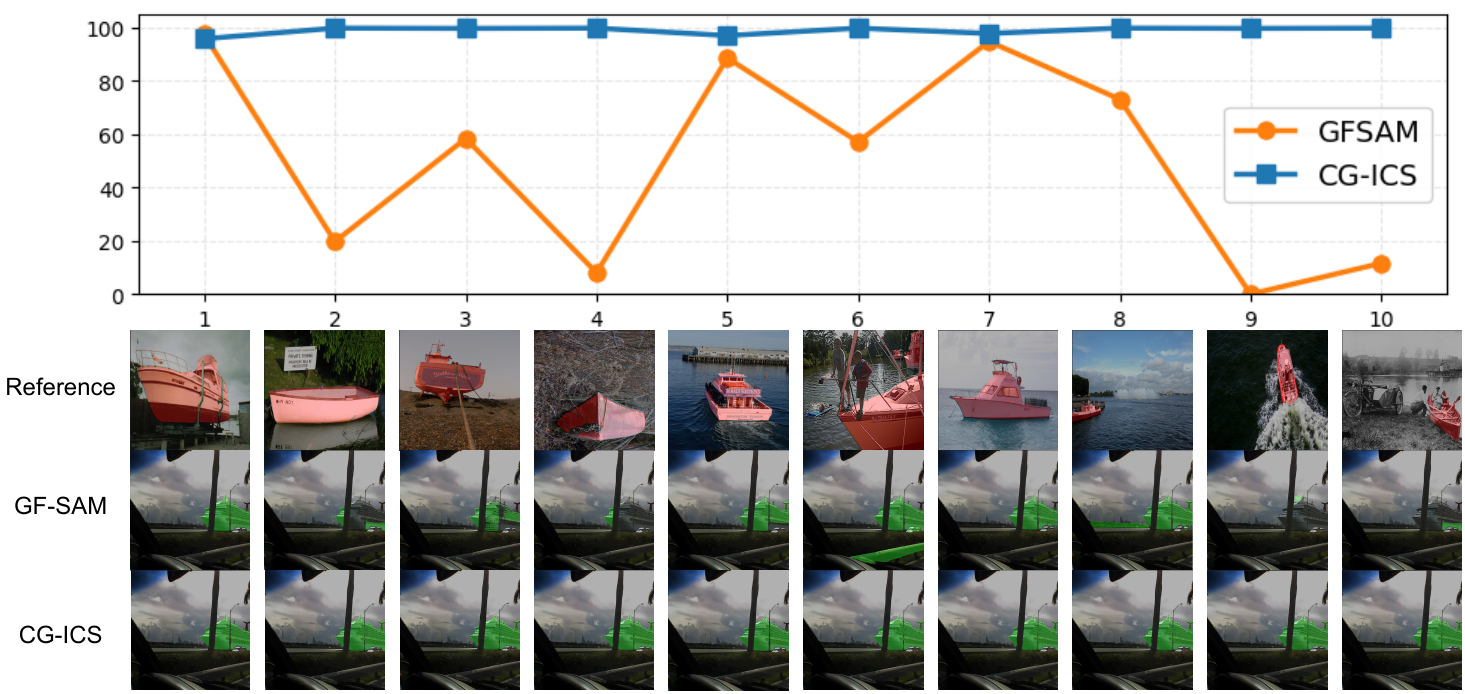}
\caption{We present the qualitative results of a query sample with multiple distinct reference examples and report the IoU achieved by GF-SAM~\cite{zhang2024gfsam} and our CG-ICS. }
\label{fig:intro_sen_com}
\end{figure}

While recent ICS methods have made notable progress in segmentation accuracy~\cite{seggpt23,liu2024simple,meng2024segic,persam24,liu2023matcher,zhang2024gfsam}, they largely overlook an equally critical aspect: system robustness, i.e., whether the model can produce stable segmentation results for the same query under different references. 
Due to the few-shot nature of ICS, where predictions are guided by only a handful of in-context examples, they are highly sensitive to the choice of reference instances.
To make this issue concrete, we probe a state-of-the-art training-free method, GF-SAM~\cite{zhang2024gfsam}, by varying the in-context references while keeping the query fixed.
As shown in Fig.~\ref{fig:intro_sen_com}, GF-SAM exhibits pronounced variance across different references and can even fail catastrophically when paired with unfavorable reference instances. 
Recently, several works~\cite{zhang2023makes,xu2024towards,suo2024rethinking} have also exposed this sensitivity, but they mainly focus on selecting the best prompt from an available pool of candidate references. 
However, such a candidate pool is often absent in real deployments, where users provide arbitrary references on the fly, and the system has no opportunity to search for a better alternative. 
UNICL-SAM~\cite{sheng2025unicl} also recognizes the robustness issue and attempts to incorporate the uncertainty of reference samples into the training process. However, it only focuses on the model’s performance when the reference samples are corrupted, without considering references that are inherently low quality. 
Therefore, a fundamental requirement emerges: an ICS system should remain accurate and stable under any reasonable reference, rather than performing well only when paired with a fortunate one.

In this paper, we take robustness as a first-class objective for ICS, and aim to reduce the prediction variance induced by reference choice while preserving strong segmentation quality.   
Our methodology is directly inspired by the rapid evolution of segmentation foundation models~\cite{sam1,ravi2024sam2,carion2025sam3segmentconcepts}. 
In particular, SAM3~\cite{carion2025sam3segmentconcepts} introduces a Promptable Concept Segmentation (PCS) paradigm, where the model can effectively interpret a textual concept prompt and produce accurate segmentation masks accordingly. 
We posit that such a concept-guided formulation offers a promising direction for advancing ICS, as textual concepts provide a higher-level and more invariant representation of the target than brittle low-level visual correspondences. 
Consequently, concept-based guidance is expected to yield more robust and stable predictions under varying references. 

However, adapting PCS to the ICS task presents a unique challenge: the standard ICS setting strictly provides only a reference image and a binary mask, lacking any prior category names or textual descriptions.  
In that case, our primary objective is to autonomously derive these concepts.  
The most straightforward solution is to leverage a model that can understand images and output textual descriptions.  
Therefore, we adopt a Multimodal Large Language Model (MLLM)~\cite{Qwen-VL,Qwen2-VL,Qwen2.5-VL,Qwen3-VL,team2023gemini}, as the concept generator, which not only meets the above requirement but also exhibits strong open-world generalization capabilities to resolve visual ambiguities and improve robustness.  
Since the MLLM and SAM3 are decoupled, a key challenge lies in selecting the optimal concept that can best prompt SAM3 for accurate segmentation.  
Therefore, as illustrated in Fig.~\ref{fig:framework}, we design a tree-search-based Concept Reasoning procedure.  
Specifically, the MLLM-driven Concept Generation module expands tree branches by producing multiple candidate concepts as tree nodes.  
For each node, a SAM3-based Concept Scoring module evaluates the candidate by jointly considering the reference and the query.  
The resulting scores determine whether the current concept should be expanded (to generate refined child nodes) or pruned.  
This iterative expand--score--prune process continues until a predefined stopping criterion is met, yielding the best concept for prompting SAM3.  
However, relying solely on linguistic concepts remains insufficient in challenging cases.  
SAM3 also highlights that providing visual exemplars (bounding boxes of the target) can help specify uncommon categories~\cite{carion2025sam3segmentconcepts}.
Therefore, we introduce a parallel Visual Exemplars Extraction route that extracts query-side visual exemplars from the concatenated reference-query image.
Finally, CG-ICS fuses the selected textual concept with the visual exemplars in a single SAM3 inference to produce the query mask.  
We term this framework \textbf{Concept-Guided In-Context Segmentation (CG-ICS)}.  
Overall, CG-ICS effectively recasts ICS as a PCS problem, leading to not only improved segmentation accuracy but also substantially enhanced robustness to reference selection, as evidenced in Fig.~\ref{fig:intro_sen_com}.  

Our contributions and the main findings are summarized as follows:  
\begin{itemize}  
    \item We tackle a critical yet underexplored issue in ICS, namely the pronounced sensitivity to in-context references. To address this problem, we make the first attempt to build a robust ICS system that maintains stable performance under diverse and low-quality references, rather than focusing on selecting an optimal reference as in prior work.  
    \item The proposed CG-ICS introduces a concept reasoning module to extract the most suitable textual concept, together with a parallel visual exemplar route that provides query-side spatial grounding. With this design, we, for the first time, successfully bring the powerful SAM3 into the ICS.  
    \item Through extensive experiments on four ICS benchmarks under both standard and robust evaluation protocols, we demonstrate that CG-ICS achieves strong segmentation performance while substantially reducing variance across diverse reference selections and quality, offering practical insights for building more reliable ICS systems.  
\end{itemize}

%% file: content/sec/2_related.tex
\section{Related work}
\paragraph{In-Context Segmentation.}
In-context segmentation (ICS) has emerged as a practical variant of visual in-context learning, where a model segments the target region in a query image conditioned on a few reference image--mask pairs without updating parameters.  
Since Bar et al.~\cite{bar2022visual} first introduced the in-context learning concept into vision, this paradigm has rapidly attracted extensive attention and led to a series of generalist segmentation systems.  
Existing ICS methods can be broadly categorized into training-based and training-free paradigms.  
Training-based methods explicitly learn the ICS paradigm by jointly feeding the in-context reference and the query into a unified model during training.  
Painter~\cite{painter23} and SegGPT~\cite{seggpt23} formulate ICS via masked image modeling~\cite{he2022masked} on the concatenated image of reference and query.  
Another line of works, including SegIC~\cite{meng2024segic}, SINE~\cite{liu2024simple}, and Iris~\cite{gao2025show}, freeze a powerful vision foundation model~\cite{caron2021emerging,dinov223} to extract dense correspondences and train a lightweight decoder to predict masks.  
Diffusion-based~\cite{ho2020denoising,rombach2022high} approaches such as DiffewS~\cite{zhu2024unleashing} and LDIS~\cite{wang2025explore} explore generative modeling as a vehicle for in-context segmentation.  
In addition, some methods adapt promptable segmenters by fine-tuning the SAM~\cite{sam1,ravi2024sam2} family for ICS, such as VRP-SAM~\cite{sun2024vrp} and SANSA~\cite{cuttano2025sansa}.
Despite strong performance, training-based ICS typically requires substantial in-domain training data and may exhibit limited generalization when faced with novel domains, corrupted annotations, or distribution shifts.  
Training-free methods avoid parameter updates by extracting visual correspondences between reference and query and converting them into prompts for a frozen promptable segmenter (e.g., SAM).  
PerSAM~\cite{persam24}, Matcher~\cite{liu2023matcher}, and GF-SAM~\cite{zhang2024gfsam} follow this paradigm by deriving point or region cues from reference-query matching to drive mask prediction.  
While these approaches primarily focus on improving average segmentation accuracy, the robustness of ICS systems under diverse and imperfect reference choices remains underexplored.  
A few studies~\cite{zhang2023makes,suo2024rethinking} have noticed the sensitivity of ICS to references, but they mainly aim to retrieve the best one from an available pool rather than ensuring stable behavior under arbitrary references.  
In contrast, our work targets high performance and low variance across diverse reference examples, which better matches real-world usage where reference quality and choice cannot be guaranteed. 

\paragraph{Robust Segmentation.}
Robust segmentation~\cite{endo2023semantic, guo2019degraded, rajagopalan2023improving, xia2019cooperative, yang2022self,chen2024robustsam} studies examine how dense prediction models behave under distribution shifts and real-world degradations. 
More recently, robustness has been explored in few-shot segmentation and in-context segmentation settings, where the prediction can be highly sensitive to the choice and quality of reference examples.  
Tang \emph{et al.}~\cite{tang2025overcomingsupportdilutionrobust} estimate the contribution of each reference image and then enhance or prune low-utility references to improve few-shot segmentation performance.  
UNICL-SAM~\cite{sheng2025unicl} reduces the performance drop under corrupted references by injecting uncertainty signals during training, which makes the model less brittle to degraded references.  
However, it does not explicitly evaluate stability across different references, and its training-based paradigm also incurs higher data and computation requirements.  
In contrast, our method adopts a training-free design that leverages advanced foundation models to perform ICS, offering a more flexible and lightweight solution while directly targeting robustness to reference variation.

%% file: content/sec/3_method.tex
\section{Method}
We present Concept-Guided In-Context Segmentation (CG-ICS), a framework that performs ICS without updating any model parameters.  
Built on MLLM and SAM3, CG-ICS derives textual concepts and visual exemplars and combines them as prompts for final mask prediction.  
We first introduce the ICS problem setup and the two foundation models in preliminaries, and then present CG-ICS in detail.  The pipeline is described in a one-shot setting for clarity, and then extended to multiple reference examples.  

\subsection{Preliminaries}

\paragraph{ICS.}
Given a single user-provided reference example $(I_r, M_r)$ and a query image $I_q$, ICS aims to produce a mask $M_q$ that captures the content in the query image.  
Only the reference image and mask are provided, and no category label or predefined class information is assumed.  

\paragraph{MLLM.}
A MLLM is a generative model that jointly processes visual inputs and natural language, enabling unified perception and language reasoning.  
MLLMs are known to be robust in handling diverse scenes and open-vocabulary descriptions, and can generalize well under distribution shifts and noisy visual conditions.  
Formally, given an image (or a set of images) and a textual instruction prompt, an MLLM produces a textual response as  
\begin{equation}
T = \mathcal{M}_{\text{MLLM}}\!\left(I, P\right),
\end{equation}
where $\mathbf{I}$ denotes the visual input, $\mathbf{P}$ denotes the input prompt, and $\mathbf{T}$ is the generated text output.  

\paragraph{SAM3.}
SAM3 is a foundation model that performs promptable concept segmentation.
The target can be specified by a short noun phrase (Textual Concept), a visual exemplar bounding box, or both.  
Given a concept prompt, SAM3 predicts semantic and instance-level masks for all instances matching the concept.  
The model also includes a presence head to indicate whether the queried concept is present in the input image.  
Formally, we write SAM3 inference as  
\begin{equation}
\Big(M^{\mathrm{ins}},\, M^{\mathrm{sem}},\, S^{\mathrm{pres}}\Big)
= \mathcal{M}_{\mathrm{SAM3}}\!\left(I,\, T,\, V\right),
\end{equation}
where $I$ is the image, $T$ is a textual concept prompt, $V$ is the visual exemplar boxes, and the outputs are instance masks $M^{\mathrm{ins}}$, a semantic mask $M^{\mathrm{sem}}$, and a presence score $S^{\mathrm{pres}}$.  

\begin{figure}[t]
  \begin{center}
  \centering
  
  \includegraphics[width=1\textwidth]{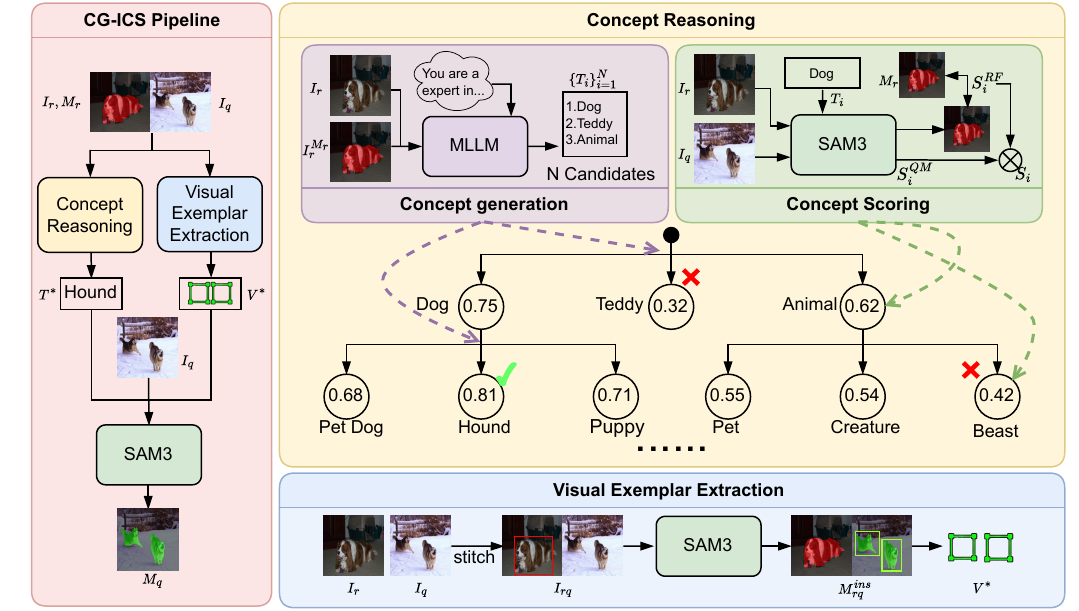}
    
      \caption{Overview of the proposed CG-ICS framework. Given a reference image and mask $(I_r, M_r)$ together with a query image $I_q$, CG-ICS performs concept reasoning and visual exemplar extraction in parallel. The concept reasoning formulates concept selection as a score-guided tree search, where the MLLM performs concept generation to expand branches and SAM3 serves as the scoring function to evaluate nodes, yielding the best-scoring concept for the target. Meanwhile, the visual exemplar extraction uses SAM3 to obtain a visual prior from the reference--query pair. Finally, SAM3 fuses the selected concept with the visual exemplars to predict the target mask.}
 \label{fig:framework}
  \end{center}
\end{figure}

\subsection{Concept Reasoning}
We propose a tree-search-based reasoning strategy to automatically extract the most suitable concept for SAM3 from the in-context references. 
The problem is formulated as searching over a concept tree, where each node corresponds to a candidate textual concept and each edge denotes a refinement from a parent concept to a more specific one.  

\paragraph{Concept Generation.}
We extract candidate textual concepts from the reference image--mask pair using a frozen MLLM, which acts as the branch expansion operator in our concept tree.  
To preserve global scene context while highlighting target details, we provide the MLLM with a two-view visual input.  
The first view is the original reference image $I_r$.  
The second view is a highlighted image $I_r^{\mathrm{M_r}}$, obtained by overlaying the masked region in red on $I_r$.  
This two-view design retains full-image context from $I_r$ while enforcing fine-grained understanding of the target object, producing more discriminative branches for subsequent search.  
The MLLM was then queried with an instruction prompt $P_{\mathrm{gen}}$ that follows three rules.  
(1) It requests a short noun phrase to match the textual concept input required by SAM3.  
(2) It explicitly asks the model to focus on the red-highlighted masked region.  
(3) If a parent-node concept is provided, it instructs the model to propose either synonymous alternatives or more fine-grained refinements conditioned on that concept, enabling progressive expansion along the tree.  
Formally, we generate the candidates set as  
\begin{equation}
\{T_i\}_{i=1}^{N} = \mathcal{M}_{\mathrm{MLLM}}\!\left(I_r,\, I_r^{\mathrm{M_r}},\, P_{\mathrm{gen}}\right),
\end{equation}
\label{concept-gen}
where $\{T_i\}_{i=1}^{N}$ denotes the resulting concept candidates.

\paragraph{Concept Scoring.}
Given a candidate concept $T_i$, we design a SAM3-based scoring function that serves as the \emph{node evaluation} module in our concept tree.  
This score quantifies how well $T_i$ (i) explains the annotated reference target and (ii) transfers to the query image. 
We first define Reference Fidelity (RF) score to measure whether the concept accurately describes the reference target indicated by $M_r$.
For each $T_i$, we prompt it to SAM3 on the reference image $I_r$ and obtain the semantic mask $M^{\mathrm{sem}}_{r,i}$:
\begin{equation}
M^{\mathrm{sem}}_{r,i} = \mathcal{M}_{\mathrm{SAM3}}\!\left(I_r,\, T_i,\, \varnothing\right)^{\mathrm{sem}}.
\end{equation}
It was then compared with the ground-truth reference mask $M_r$ by computing their IoU. We use the IoU as the RF score:
\begin{equation}
S^{\mathrm{RF}}_i = \mathrm{IoU}\!\left(M^{\mathrm{sem}}_{r,i},\, M_r\right).
\end{equation}
Besides, the concept should also match the to-be-segmented part in the query image.  
We prompt SAM3 with the same $T_i$ on the query image $I_q$ and use the presence score as the Query Matchability (QM) score:
\begin{equation}
S^{\mathrm{QM}}_i = \mathcal{M}_{\mathrm{SAM3}}\!\left(I_q,\, T_i,\, \varnothing\right)^{\mathrm{pres}}.
\end{equation}
Finally, we combine the two criteria with a multiplicative rule to obtain the concept score:
\begin{equation}
S_i = S^{\mathrm{RF}}_i \cdot S^{\mathrm{QM}}_i.
\end{equation}
\label{score}

\paragraph{Tree Search for Concept Refinement.}
Based on Concept Generation and Concept Scoring, we perform a score-guided tree search to refine textual concepts.  
We maintain a frontier $\mathcal{F}^{(k)}$ at round $k$, and initialize the search by expanding the root node with Concept Generation to obtain $N$ first-layer candidates.  
For each node concept $T_i^k\in\mathcal{F}^{(k)}$, we compute its score $S_i$ using Concept Scoring.  
We prune nodes with $S_i^k<\tau_{\mathrm{pruned}}$, and only expand the remaining ones.  
Each surviving concept node will be passed into Concept Generation again to produce $N$ children concepts, including near-synonyms and finer-grained refinements.
We maintain a concept buffer $\mathcal{B}$ that stores all visited concepts to prevent duplicate generation during expansion. 
The resulting children form the next frontier $\mathcal{F}^{(k+1)}$, and the expand--score--prune loop repeats up to $K$ rounds.  
The looping will be early stopped if any node reaches $S_i^k\ge \tau_{\mathrm{stop}}$. If all nodes in a round are fully pruned, the expansion will restart by returning to the parent node to generate totally different concepts.
Finally, we select the best concept with the highest score in $\mathcal{B}$ by
\begin{equation}
T^{*}=\arg\max_{\mathcal{B}} S_i.
\end{equation}

\subsection{Visual Exemplar Extraction}
Relying on textual concepts alone can be unreliable, since MLLMs may hallucinate or output over-specific phrases that do not correspond to the true target.
Moreover, SAM3 highlights that providing visual exemplars (bounding boxes of the target) can help specify uncommon categories.
However, the standard SAM3 exemplar prompt is defined within a single image and does not directly support cross-image exemplar transfer. 
Inspired by SegGPT~\cite{seggpt23}, which stitches the reference and query in a single input, we propose a simple yet effective way to derive a visual exemplar of the query image.

We first form a stitched image by concatenating the reference and query images horizontally:   
\begin{equation}
I_{rq} = \mathrm{Stitch}(I_r,\, I_q).
\end{equation}
The reference exemplar boxes can be extracted from the reference mask.  
\begin{equation}
V_r = \mathrm{BBox}(M_r).
\end{equation}
Next, we use $V_r$ as a visual exemplars prompt to invoke SAM3 on the stitched image.  
\begin{equation}
M^{\mathrm{ins}}_{rq} = \mathcal{M}_{\mathrm{SAM3}}\!\left(I_{\mathrm{rq}},\, \varnothing,\, V_r\right)_\mathrm{ins}.
\end{equation}
Since SAM3 can segment all matched instances under a prompt, it can return masks on both the reference side and the query side within $I_{\mathrm{rq}}$. 
We take the instance mask on the query half, denoted as $M^{\mathrm{ins}}_{q} = \mathrm{RightHalf}(M^{\mathrm{ins}}_{rq})$, and convert it into coarse target exemplar boxes:  
\begin{equation}
V^{*} = V_q = \mathrm{BBox}\!\left(M^{\mathrm{ins}}_{q}\right).
\end{equation}
The resulting $V^*$ serves as the visual exemplars, which can be subsequently used together with the textual concept to stabilize the final segmentation.  

\subsection{Segmentation with Joint Prompts}
After selecting the best textual concept $T^{*}$ and deriving the target exemplar boxes $V^*$, we feed them jointly to SAM3 for the final segmentation.  
The textual concept provides category-level semantics, while the visual exemplars constrain the spatial reference on the query image.  
Formally, we perform the final inference on the query image as  
\begin{equation}
\Big(M^{\mathrm{ins}}_q,\, M^{\mathrm{sem}}_q,\, S^{\mathrm{pres}}_q\Big)
= \mathcal{M}_{\mathrm{SAM3}}\!\left(I_q,\, T^{*},\, V^{*}\right).
\end{equation}
We take the semantic mask $M^{\mathrm{sem}}_q$ as the final prediction ${M}_q$.

\subsection{Multiple-Reference Setting}
When multiple reference examples $\{(I_r^{n}, M_r^{n})\}_{n=1}^{m}$ are available, we extend CG-ICS to the multiple-shot setting.  
For textual concepts, due to the MLLM's limited ability to reliably reason over many images at once, we perform concept reasoning on each reference independently and obtain one final concept per reference, denoted as $\{T^{*}\}_{n=1}^{m}$.  
We then compare these $m$ concepts and select the best one by re-scoring each of them on all reference image-mask pairs to measure cross-reference consistency.   
For visual exemplar extraction, we stitch each reference with the query separately to obtain query-side bboxes from each reference, and then collect all such bboxes as visual exemplars.

%% file: content/sec/4_experiments.tex
\section{Experiments}
\subsection{Datasets and Protocol}
\paragraph{Datasets.}
To demonstrate semantic ICS capability, we evaluate CG-ICS on four representative ICS benchmarks, including Pascal-5$^{i}$, COCO-20$^{i}$, LVIS-92$^{i}$, and FSS-1000.  
\textbf{Pascal-5$^{i}$} is built upon Pascal VOC 2012~\cite{pascal10} and SDS~\cite{sds11}.  
It contains 20 object categories that are split into 4 folds, with 5 classes per fold.  
\textbf{COCO-20$^{i}$} is derived from MS COCO~\cite{coco14} and includes 80 categories in total. The 80 classes are partitioned into 4 folds, with 20 classes per fold.  
\textbf{LVIS-92$^{i}$}~\cite{liu2023matcher} is built based on LVIS~\cite{gupta2019lvis} which is more challenging. It selects 920 categories with more than two images per class and divides them into 10 folds.
\textbf{FSS-1000}~\cite{fss1000} contains 1000 categories that are split into training, validation, and test containing 520, 240, and 240 categories, respectively.  

\paragraph{Protocol.}
Three types of experiments are conducted. 
\textbf{(1) Standard ICS} evaluates overall segmentation performance as in prior ICS work, and thus we follow the setting of Matcher~\cite{liu2023matcher} for fair comparison. 
For Pascal-5$^{i}$ and COCO-20$^{i}$, we sample 1{,}000 queries per fold from the validation split, for LVIS-92$^{i}$ we sample 2{,}300 queries per fold, and for FSS-1000 we use all queries in the official test split.  
We evaluate both 1-shot and 5-shot settings on all datasets and report the mean Intersection-over-Union \emph{(mIoU)} as the metric. 
\textbf{(2) Robustness to Reference Selection.}
For each fold on Pascal-5$^{i}$, COCO-20$^{i}$, and LVIS-92$^{i}$, we sample 500 query images and pair each query with 50 different reference examples from the same class.  
In addition to the mean IoU, the Standard Deviation \emph{(Std)} and the Coefficient of Variation $CV=\frac{Std}{mIoU}$ across these references are also reported. The lower values of \emph{Std} and \emph{CV} indicate better relative stability. 
\textbf{(3) Robustness to Corrupted References.}  
Following UNICL-SAM~\cite{sheng2025unicl}, we apply diverse corruptions to the reference examples and  
report the $mIoU$ with clean data as well as the performance drop under corruptions.  
This evaluation is conducted on 1500 sampled instances from COCO-20$^{i}$.  

\subsection{Implementation Details.}
We adopt Qwen3-VL-4B~\cite{Qwen3-VL} as the MLLM backbone for concept generation.  
For scoring and mask prediction, the official SAM3~\cite{carion2025sam3segmentconcepts} release is adopted as the promptable segmentation backbone.  
All model parameters are frozen.  
For the tree search, each successful node expands to $N{=}5$ refined concept candidates generated by the MLLM.  
A node is pruned if its score is smaller than $\tau_{\mathrm{pruned}}{=}0.5$.  
We enable early stopping when any node reaches $\tau_{\mathrm{stop}}{=}0.8$ to reduce computation.  
The search runs for at most $K{=}3$ refinement rounds.

\begin{table*}[t]
\centering
\caption{
Comparison with state-of-the-art methods under the \textbf{Standard ICS} protocol. 
\graytext{Gray} indicates methods trained on in-domain datasets that include the test categories. 
\textbf{Bold} denotes the best result, and \underline{underline} denotes the second best. 
}
\label{tab:sota}
\setlength{\tabcolsep}{4pt}
\begin{tabular}{l cccccccc}
\toprule
\multirow{2}{*}{Methods} &
\multicolumn{2}{c}{Pascal-5$^i$} &
\multicolumn{2}{c}{COCO-20$^i$} &
\multicolumn{2}{c}{LVIS-92$^i$} &
\multicolumn{2}{c}{FSS-1000} \\
\cmidrule(lr){2-3}\cmidrule(lr){4-5}\cmidrule(lr){6-7}\cmidrule(lr){8-9}
& 1-shot & 5-shot & 1-shot & 5-shot & 1-shot & 5-shot & 1-shot & 5-shot \\
\midrule
\rowcolor[gray]{.8}\multicolumn{9}{c}{\emph{Training-based}}\\
SegGPT~\cite{seggpt23}                & \graytext{83.0} & \textbf{\graytext{89.8}} & \graytext{56.1} & \graytext{67.9} & 18.6 & 25.4 & 85.6 & 89.3  \\
SINE~\cite{liu2024simple}             & 85.4 & 86.2 & \graytext{64.5} & \graytext{66.1} & 31.2 & 35.5 &  --  &  --   \\
DiffwS~\cite{zhu2024unleashing}       & \underline{\graytext{88.3}} & \graytext{87.8} & \graytext{71.3} & \graytext{72.2} & 31.4 & 35.4 & 87.8 & 88.0  \\
UNICL-SAM~\cite{sheng2025unicl} &-&-&\textbf{\graytext{77.8}} & \textbf{\graytext{78.7}} & 34.1 & 37.1 & 84.0 & 86.3 \\
SANSA~\cite{cuttano2025sansa}         & -& -    & \underline{\graytext{75.6}} & \underline{\graytext{78.6}} & \underline{\graytext{50.3}} & \graytext{\textbf{59.0}} & \underline{90.0} & \textbf{91.0} \\
\midrule
\rowcolor[gray]{.8}\multicolumn{9}{c}{\emph{Training-free}}\\
Matcher~\cite{liu2023matcher}              & 68.1 & 74.0 & 52.7 & 60.7 & 33.0 & 40.0 & 87.0 & 89.6  \\
GF-SAM~\cite{zhang2024gfsam}         & 72.1 & 82.6 & 58.7 & 66.8 & 35.2 & 44.2 & 88.0 & 88.9  \\
CG-ICS (ours) & \textbf{89.3} &  \underline{89.6}  & 72.3 & 72.6  &  \textbf{55.4}  &  \underline{56.2}  & \textbf{90.2} &   89.3  \\
\bottomrule
\end{tabular}
\end{table*}

\subsection{Comparison with state-of-the-art.}

\paragraph{Standard ICS.}
We compare CG-ICS with state-of-the-art training-free and training-based generalist ICS methods in Table~\ref{tab:sota}.
Among training-free approaches, CG-ICS delivers a clear and consistent lead across all benchmarks, e.g., $+17.2$ $mIoU$ on Pascal-5$^{i}$ , $+13.6$ on COCO-20$^{i}$ , and $+20.2$ on LVIS-92$^{i}$ in 1-shot setting over the strongest training-free baseline GF-SAM~\cite{zhang2024gfsam}.
Notably, despite being training-free, CG-ICS remains competitive with training-based methods trained with in-domain data that includes the test categories.
CG-ICS still achieves state-of-the-art results on Pascal-5$^{i}$ ($89.3$), LVIS-92$^{i}$ ($55.4$), and FSS-1000 ($90.2$) in 1-shot setting.
In the remaining settings, CG-ICS only slightly trails the best training-based competitors, e.g., $0.2$ $mIoU$ on Pascal-5$^{i}$ 5-shot and $3.3/6.0$ $mIoU$ on COCO-20$^{i}$ 1-shot/5-shot.
Overall, strong performance across diverse datasets indicates that our CG-ICS is a powerful paradigm.

\begin{table*}[t]
\centering
\caption{Comparison results under the Robustness to Reference Selection protocol. \graytext{Gray} indicates methods trained on in-domain datasets that include the test categories. 
\textbf{Bold} denotes the best result, and \underline{underline} denotes the second best. }
\label{tab:robust_ics_with_cv}

\begin{tabular}{lccccccccc}
\toprule
\multirow{2}{*}{Methods} &
\multicolumn{3}{c}{Pascal-5$^i$} &
\multicolumn{3}{c}{COCO-20$^i$} &
\multicolumn{3}{c}{LVIS-92$^i$} \\
\cmidrule(lr){2-4}\cmidrule(lr){5-7}\cmidrule(lr){8-10}
& $mIoU$ & $Std$ & $CV$ & $mIoU$ & $Std$ & $CV$ & $mIoU$ & $Std$ & $CV$ \\
\midrule

\rowcolor[gray]{.8}\multicolumn{10}{c}{\emph{Training-based}}\\
SegGPT~\cite{seggpt23}            & \graytext{75.1} & \graytext{17.3} & \graytext{23.0\%} & \graytext{45.7} & \graytext{22.7} & \graytext{49.7\%} & 19.4 & \textbf{13.5} & 69.6\%\\
SINE~\cite{liu2024simple}         & \graytext{81.7} & \graytext{15.8} & \graytext{19.3\%} & \graytext{68.3} & \graytext{16.1} & \graytext{23.6\%} & 30.9 & 20.3 & 65.6\%\\
SANSA~\cite{cuttano2025sansa}     & \underline{83.9} &  \underline{9.3} & \underline{11.1\%} & \graytext{\textbf{74.3}} & \graytext{\underline{13.1}} & \graytext{\underline{17.6\%}} & \graytext{\underline{50.8}} & \graytext{18.2} & \graytext{\underline{35.8\%}}\\
\midrule

\rowcolor[gray]{.8}\multicolumn{10}{c}{\emph{Training-free}}\\
Matcher~\cite{liu2023matcher}     & 65.3 & 20.6 & 31.5\% & 53.9 & 18.7 & 34.7\% & 38.8 & 19.2 & 49.4\%\\
GF-SAM~\cite{zhang2024gfsam}      & 70.6 & 19.3 & 27.3\% & 58.3 & 16.4 & 28.1\% & 42.2 & 18.7 & 44.3\%\\
CG-ICS(ours)                      & \textbf{88.4} &  \textbf{6.9} &  \textbf{7.8}\% & \underline{72.1} & \textbf{9.3} & \textbf{12.9\%} & \textbf{57.1} & \underline{17.2} & \textbf{30.1\%}\\
\bottomrule
\end{tabular}
\end{table*}

\paragraph{Robustness to Reference Selection.}
Table~\ref{tab:robust_ics_with_cv} reports robustness under varying in-context references.  
Across all three benchmarks, CG-ICS obtains the lowest $CV$, indicating the most stable segmentation relative to its mean performance.  
On Pascal-5$^{i}$ and COCO-20$^{i}$, CG-ICS achieves the lowest $CV$ ($7.8\%$ and $12.9\%$), markedly improving over GF-SAM ($27.3\%$ and $28.1\%$) and remaining more stable than training-based baselines such as SANSA ($11.1\%$ and $17.6\%$).  
On LVIS-92$^{i}$, CG-ICS still yields the best robustness with $30.1\%$ $CV$, outperforming SANSA ($35.8\%$) and largely reducing the instability of SegGPT/SINE ($69.6\%/65.6\%$).  
Overall, the consistently smallest $CV$ indicates that CG-ICS maintains strong accuracy while being the least sensitive to reference selection.

\begin{table}[t]
\centering
\caption{Comparison results under the robustness to Corrupted References protocol. \graytext{Gray} indicates methods trained on in-domain datasets that include the test categories. 
\textbf{Bold} denotes the best result, and \underline{underline} denotes the second best. }
\label{tab:degrd}
\renewcommand{\arraystretch}{1.15}
\resizebox{\linewidth}{!}{
\begin{tabular}{l |c |ccccc c |c c}
\toprule
{Methods} &
{Clean} &
Color & Blurriness & Compression & Space & Domain & Deformation&
Mean &
Ratio \\
\midrule
\rowcolor[gray]{.8}\multicolumn{10}{c}{\emph{Training-based}}\\
SegGPT~\cite{seggpt23}   & \graytext{62.1} & \graytext{-2.7} & \graytext{-3.0} & \graytext{\textbf{-2.7}} & \graytext{-7.8} & \graytext{-6.2} & \graytext{-8.8} & \graytext{-5.2} & \graytext{-8.4\%} \\
SINE~\cite{liu2024simple}       & \graytext{70.0} & \graytext{-0.7} & \graytext{-1.7} & \graytext{-5.9} & \graytext{\underline{-2.8}} & \graytext{-6.5} & \graytext{-12.4} & \graytext{-5.0} & \graytext{-7.1\%} \\
SANSA~\cite{cuttano2025sansa} & \graytext{70.5} & \graytext{-1.2} & \graytext{-4.0} & \graytext{-11.1} & \graytext{-5.8} & \graytext{-12.9} & \graytext{\underline{-1.4}} & \graytext{-6.1} & \graytext{-8.6\%} \\
UNICL-SAM~\cite{sheng2025unicl}     & \graytext{\textbf{79.8}} & \graytext{-1.0} & \graytext{-1.8} & \graytext{\underline{-3.0}} & \graytext{-3.3} & \graytext{\underline{-4.8}} & \graytext{-5.0} & \graytext{-3.1} & \graytext{\underline{-3.9\%}} \\
\midrule
\rowcolor[gray]{.8}\multicolumn{10}{c}{\emph{Training-free}}\\
Matcher~\cite{liu2023matcher} & 41.6 & \underline{-0.5} & -2.5 & -6.5 & -3.8 & -5.3 & -6.2 & -4.1 & -9.9\% \\
GF-SAM~\cite{zhang2024gfsam} & 61.5 & \textbf{-0.1} & \underline{-0.8} & -5.3 & \textbf{-1.9} & \textbf{-3.8} & -3.0 & \underline{-2.5} & -4.1\% \\
CG-ICS(ours) & \underline{74.6} & \textbf{-0.1} & \textbf{-0.3} & -4.2 & -3.4 & -5.2 & \textbf{-0.5} & \textbf{-2.3} & \textbf{-3.1\%} \\
\bottomrule
\end{tabular}}
\end{table}

\paragraph{Robustness to Corrupted References.}
Table~\ref{tab:degrd} compares robustness under six corruptions by reporting the mIoU on clean data and the mIoU drops. 
CG-ICS attains the best decline Ratio across all compared generalists, reaching only $-3.1\%$, which is better than the strongest training-based baseline UNICL-SAM ($-3.9\%$). 
This advantage is not achieved by sacrificing accuracy, as CG-ICS also delivers a high clean mIoU of $74.6$. 
In contrast, other training-free methods suffer noticeably larger relative drops, e.g., GF-SAM and Matcher degrade to $-4.1\%$ and $-9.9\%$, respectively, while training-based SegGPT/SINE/SANSA are even more sensitive, with Ratio ranging from $-7.1\%$ to $-8.6\%$. 
Across individual corruptions, CG-ICS remains nearly invariant to color and blur (both $\leq 0.3$ drop) and is particularly resilient to deformation ($-0.5$), which jointly contributes to its lowest overall Ratio. 
Overall, the consistently smallest Ratio demonstrates that CG-ICS provides state-of-the-art robustness against distribution shifts, even outperforming training-based generalists.

\subsection{Ablation Studies.}
We conduct ablation studies under the 1-shot setting of Robustness to Reference Selection on COCO-20$^{i}$, which are the most widely used ICS benchmarks.

\begin{table}[t]
\centering
\caption{Ablation study of the each component on COCO-20$^{i}$.  
The first row represents directly feeding a single concept prompt generated by the MLLM into SAM3 for segmentation.  
We then incrementally add candidate sampling, RF/QM scoring, tree-search, and the visual branch. }
\label{tab:scoring_ablation}

\begin{tabular}{cccc c ccc}
\toprule
\multicolumn{4}{c}{Concept Reasoning} & \multirow{2}{*}{Visual} & \multicolumn{3}{c}{COCO-20$^i$} \\
\cmidrule(lr){1-4}\cmidrule(lr){6-8}
Candidates & RF-Score & QM-Score &  Search & & $mIoU$ & $Std$ & $CV$ \\
\midrule
& & & & & 67.6 & 15.1 & 22.3\% \\
\checkmark & & & & & 66.5 & 15.3 & 23.0\%\\
\checkmark & \checkmark & & & & 69.3 & 13.9 & 20.5\%\\
\checkmark &  & \checkmark & & & 62.8 & 18.2 & 29.0\%\\
\checkmark & \checkmark & \checkmark & & & 70.1 & 11.3 & 16.1\%\\
\checkmark & \checkmark & \checkmark & \checkmark & & 71.2 & 10.5 & 14.7\%\\
\checkmark & \checkmark & \checkmark & \checkmark & \checkmark & 72.1 & 9.3 & 12.9\%\\

\bottomrule
\end{tabular}
\end{table}

\paragraph{Ablation study of key components.}
As shown in Table~\ref{tab:scoring_ablation}, directly using a single MLLM-generated concept prompt yields limited performance and robustness, with $67.6$ mIoU and $22.3\%$ $CV$.  
Introducing multiple candidates alone brings marginal changes, while adding the proposed RF-Score and QM-Score consistently improves both accuracy and stability.  
In particular, combining RF and QM reduces the relative variation to $16.1\%$ $CV$ and boosts performance to $70.1$ $mIoU$, indicating that concept selection is critical under varying references.  
The search-based selection further strengthens robustness ($14.7\%$ $CV$), and enabling the visual exemplar branch achieves the best overall trade-off, reaching $72.1$ $mIoU$  with the lowest dispersion ($9.3$ Std, $12.9\%$ $CV$).  
Overall, each component contributes incrementally, and the full design is necessary to obtain both high accuracy and reliable reference-insensitive behavior.

\begin{figure*}[t]
  \centering
  \begin{minipage}[t]{0.49\textwidth}
    \centering
    \begin{subfigure}[t]{0.48\linewidth}
      \centering
      \includegraphics[width=\linewidth]{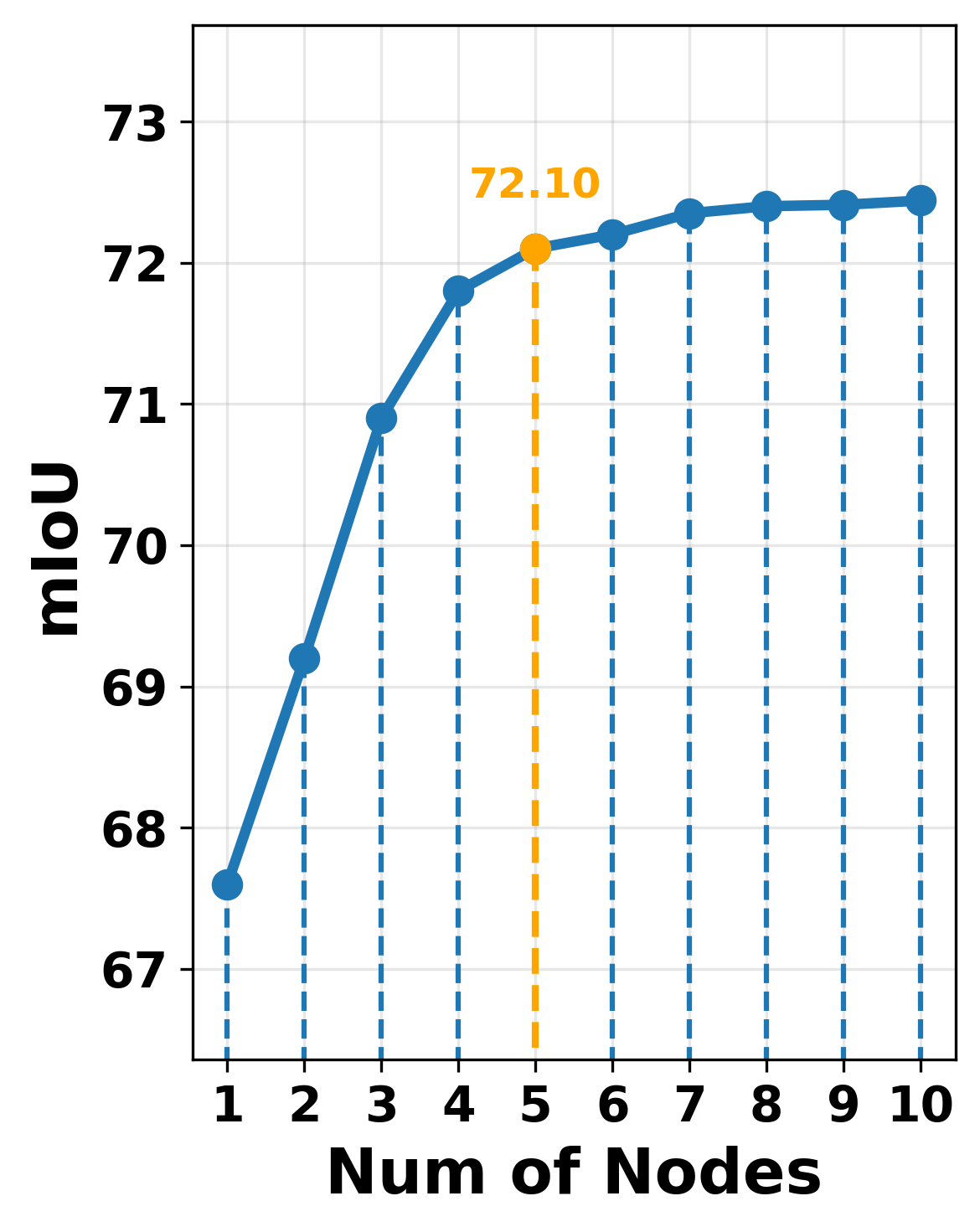}
      \caption{Performance}
      \label{fig:g1-cand-miou}
    \end{subfigure}
    \hfill
    \begin{subfigure}[t]{0.48\linewidth}
      \centering
      \includegraphics[width=\linewidth]{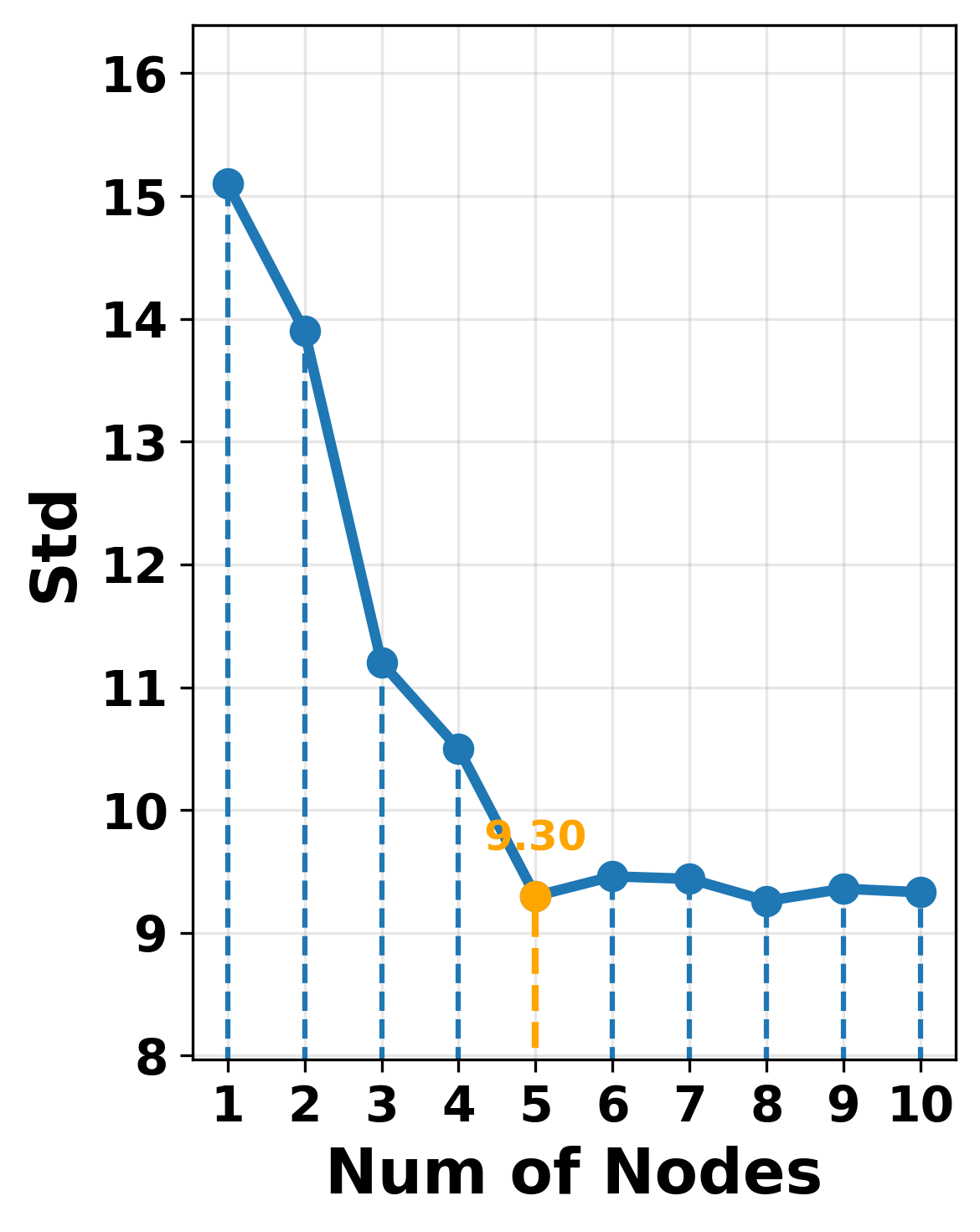}
      \caption{Robustness}
      \label{fig:g1-cand-std}
    \end{subfigure}
    \caption{Effect of the number of search nodes $N$ on performance and robustness.}
    \label{fig:g1-candidates-ablation}
  \end{minipage}
  \hfill
  \begin{minipage}[t]{0.49\textwidth}
    \centering
    \begin{subfigure}[t]{0.48\linewidth}
      \centering
      \includegraphics[width=\linewidth]{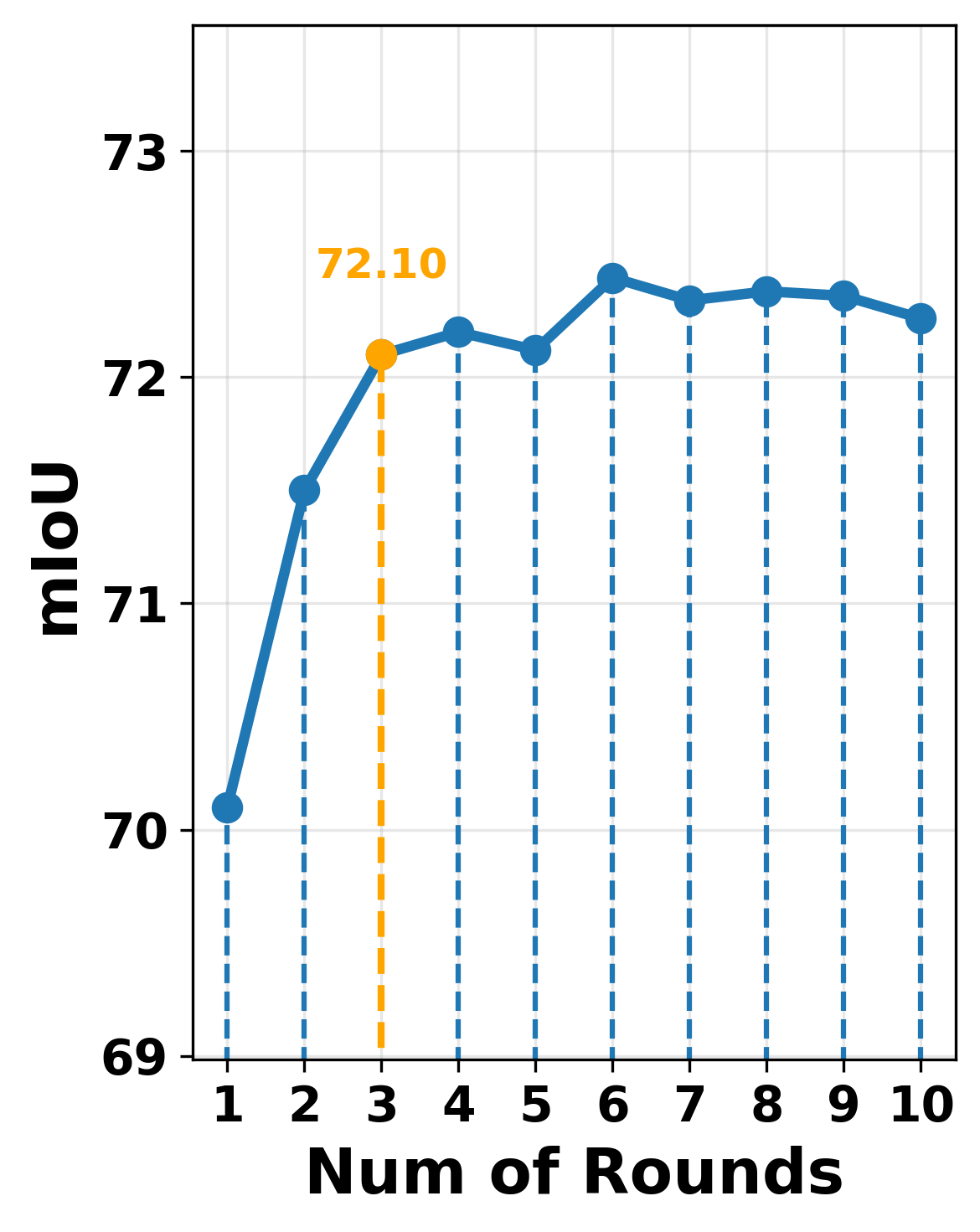}
      \caption{Performance}
      \label{fig:g2-cand-miou}
    \end{subfigure}
    \hfill
    \begin{subfigure}[t]{0.48\linewidth}
      \centering
      \includegraphics[width=\linewidth]{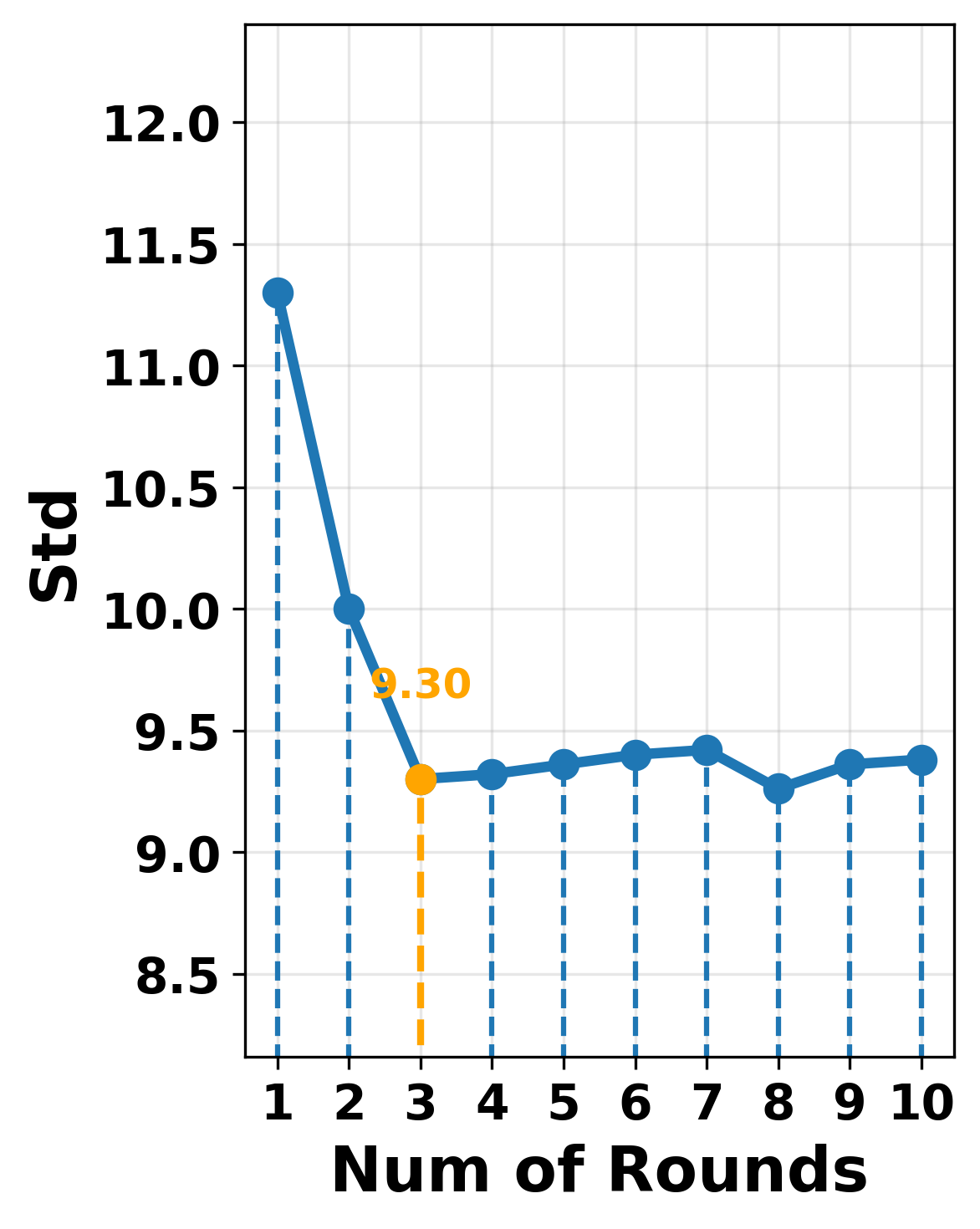}
      \caption{Robustness}
      \label{fig:g2-cand-std}
    \end{subfigure}
    \caption{Effect of the number of search rounds $K$ on performance and robustness.}
    \label{fig:g2-candidates-ablation}
  \end{minipage}
\end{figure*}

\begin{figure}[t]
\centering
\includegraphics[width=0.7\textwidth]{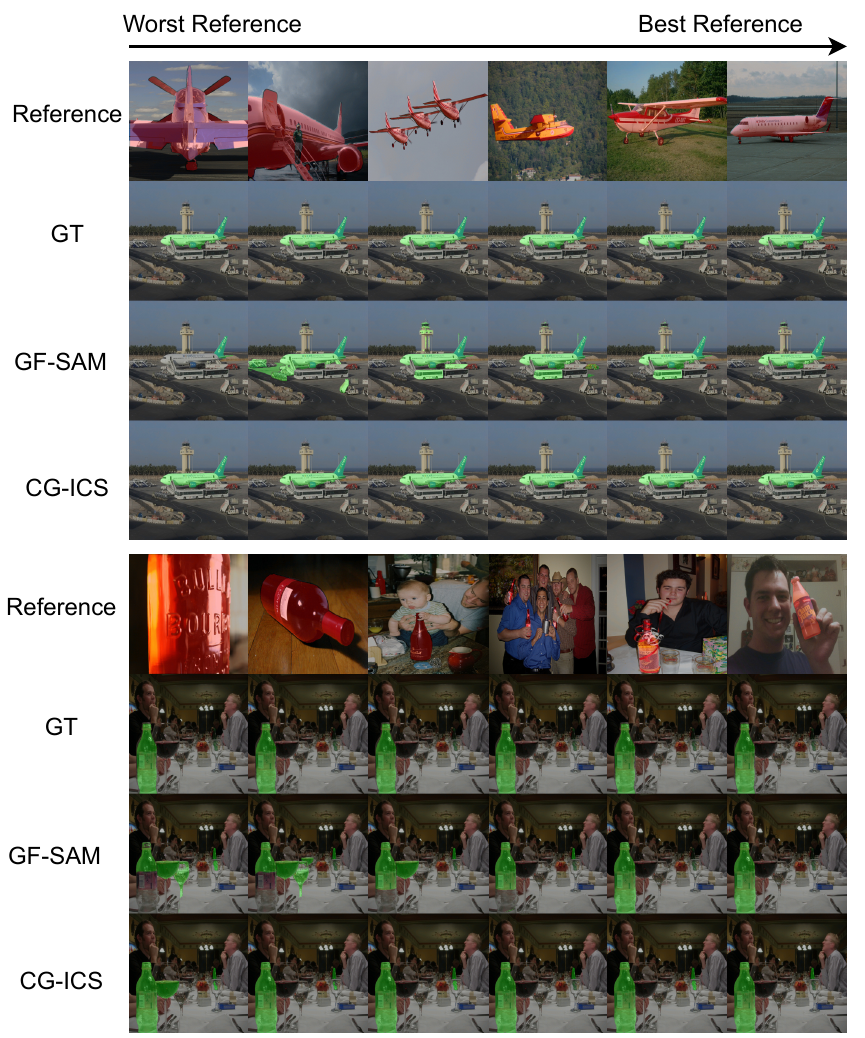}
\caption{Qualitative comparison of GF-SAM and CG-ICS under varying in-context references ranked from worst to best according to GF-SAM performance. }
\label{fig:compare_vis}
\end{figure}

\paragraph{Ablation on the search nodes and rounds.}
On COCO-20$^{i}$, increasing the number of search nodes (concept candidates) $N$ steadily improves the average segmentation accuracy and enhances robustness across different references.  
As shown in Fig.~\ref{fig:g1-cand-miou}, mIoU increases from $67.6$ at $N{=}1$ to $71.9$ at $N{=}4$, and then quickly saturates around $72.1$--$72.4$ for larger $N$ (with $72.10$ at $N{=}5$).  
Meanwhile, Fig.~\ref{fig:g1-cand-std} shows that $Std$ drops from $15.1$ at $N{=}1$ to $10.5$ at $N{=}4$, reaches $9.30$ at $N{=}5$, and stays nearly unchanged afterwards.  
Overall, both curves become almost flat once $N>5$, suggesting diminishing returns from generating more candidates.  
Therefore, we set $N{=}5$ in all experiments to balance performance, robustness, and inference cost.  
For the search rounds $K$, Fig.~\ref{fig:g2-candidates-ablation} exhibits the same trend, where both mIoU and Std change marginally once $K>3$.

\subsection{Qualitative Results.}

We visualize qualitative comparisons between GF-SAM and CG-ICS under varying in-context references in Fig.~\ref{fig:compare_vis}.
GF-SAM exhibits strong sensitivity to the reference choice, and its predictions can change noticeably as the reference varies from “worst” to “best.”
When the reference viewpoint shifts significantly (the first column of the top panel), it incorrectly segments the airplane tail instead of the whole target, indicating a tendency to match salient subparts rather than preserving the full object extent.
When the object is partially occluded (the first column in the bottom panel), it is easily distracted by occluders and co-occurring instances, and thus tends to segment the glass rather than the bottle.
Moreover, even for moderately better references, GF-SAM may still produce incomplete masks or leak to nearby regions with similar textures, suggesting inconsistent correspondence under subtle appearance changes.
This also implies that GF-SAM often requires carefully selected “good” references to avoid failure cases, limiting its practicality in real-world usage.
Such errors reflect the limitation of pure visual matching: correspondence is largely driven by local appearance similarity, which becomes unreliable under viewpoint changes and occlusions, leading to semantic drift and unstable boundaries.
In contrast, CG-ICS remains notably stable across the same reference range and consistently outputs high-quality masks, demonstrating stronger robustness to imperfect in-context references.

%% file: content/sec/5_conclusion.tex
\section{Conclusion}
\label{sec:conclusion}
In this work, we study ICS with robustness as a first-class objective and establish evaluation protocols that measure both average performance and reference-induced variance.
We introduce Concept-Guided In-Context Segmentation (CG-ICS), a training-free framework that performs concept selection through a score-guided tree search, where the MLLM performs concept generation to expand branches and SAM3 serves as the scoring function to evaluate nodes to identify a reliable textual prompt, complemented by a parallel visual exemplar route for query-side grounding.  
Across standard ICS benchmarks and robust evaluation protocols, CG-ICS achieves strong accuracy while substantially reducing variance, moving ICS toward more reliable real-world deployment.

\section*{Acknowledgements}
\label{sec:acknowledgements}
This work was supported by the New Generation Artificial Intelligence-National Science and Technology Major Project (No. 2025ZD0122701), the National Natural Science Foundation of China ( No. 62576299), and Fundamental Research Funds for the Central Universities and Xiaomi Young Talents Program.